\documentclass[10pt,twocolumn,letterpaper]{article}

\usepackage{cvpr}
\usepackage{times}
\usepackage{epsfig}
\usepackage{graphicx}
\usepackage{amsmath}
\usepackage{amssymb}
\usepackage{subfigure}

\usepackage[pagebackref=true,breaklinks=true,letterpaper=true,colorlinks,bookmarks=false]{hyperref}
\usepackage{cleveref}
\cvprfinalcopy 

\def\cvprPaperID{****} 

\ifcvprfinal\fi
\begin{document}

\title{A Simple Fix for Convolutional Neural Network via Coordinate Embedding}

\author{Liliang Ren, Zhuonan Hao\\
 University of California, San Diego \\
  La Jolla, CA92093 \\
{\tt\small \{lren,z4hao\}@ucsd.edu}\\}

\maketitle

\begin{abstract}
   Convolutional Neural Networks (CNN) has been widely applied in the realm of computer vision. However, given the fact that CNN models are translation invariant, they are not aware of the coordinate information of each pixel. Thus the generalization ability of CNN will be limited since the coordinate information is crucial for a model to learn affine transformations which directly operate on the coordinate of each pixel. In this project, we proposed a simple approach to incorporate the coordinate information to the CNN model through coordinate embedding. Our approach does not change the downstream model architecture and can be easily applied to the pre-trained models for the task like object  detection. Our experiments on the German Traffic Sign Detection Benchmark show that our approach not only significantly improve the model performance but also have better robustness with respect to the affine transformation.
\end{abstract}

\section{Introduction}
Convolutional Neural Networks (CNN) have enjoyed a wide range of applications as the most effective tool in the field of computer vision. In contrast with fully-connected neural networks, CNN exploit the image property of translation invariance for parameter sharing and thus reduce the computational complexity and improve the generalization ability.

 For the object recognition task, many one-stage CNN models that directly optimized on bounding box regression and ROI classification (e.g. Single Shot Detector (SSD)\cite{ssd} and YOLO \cite{yolo}) have been widely applied to the real world application scenarios. However, since CNN only has translation invariance and is not aware of the coordinates of each of the pixel in the image, its generalization ability will be limited when the affine distortion happens to the objects in the image. This kind of distortion commonly occurs in the object detection tasks when the objects are observed from different perspectives in the real world. 

Some detection models have taken advantage of layers to learn simple coordinate transformation, but they have been illustrated the inability with respect to spatial transformation between dense Cartesian representation and a sparse, pixel-based representation \cite{liu2018intriguing}. One feasible solution is using extra coordinate channels in convolutional layer, which is called CoordConv layer \cite{liu2018intriguing}. However, since this approach needs to change every convolutional layer to the CoordConv layer, it can not be applied to most of the pre-trained models that use vanilla convolutions, which harms the universality of this approach.

\textit{Contribution:} In this paper, we propose to incorporate the coordinate information through two learnable coordinate embeddings which are normalized and only added to each channel of the initial input image. Thus our approach does not change the downstream model architecture and can be applied to any pre-trained models in the tasks such as object object detection and recognition. Our experiments on a traffic sign detection dataset show that our approach not only significantly improve the model performance but also have better robustness with respect to the affine transformation.

\begin{figure*}[htb]
\begin{center}
\includegraphics[width=0.8\linewidth]{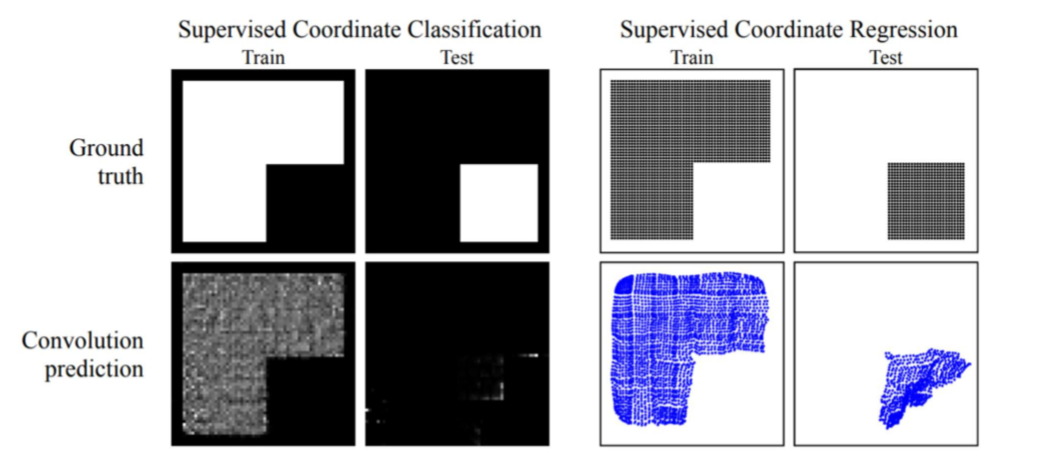}
\end{center}
   \caption{ The prediction results of Convolutional Neural Network on the Supervised Coordinate Classification and Regression tasks proposed by Liu \etal \cite{liu2018intriguing}.}
\label{fail}

\end{figure*}
\begin{figure*}[htb]
\begin{center}
\includegraphics[width=0.8\linewidth]{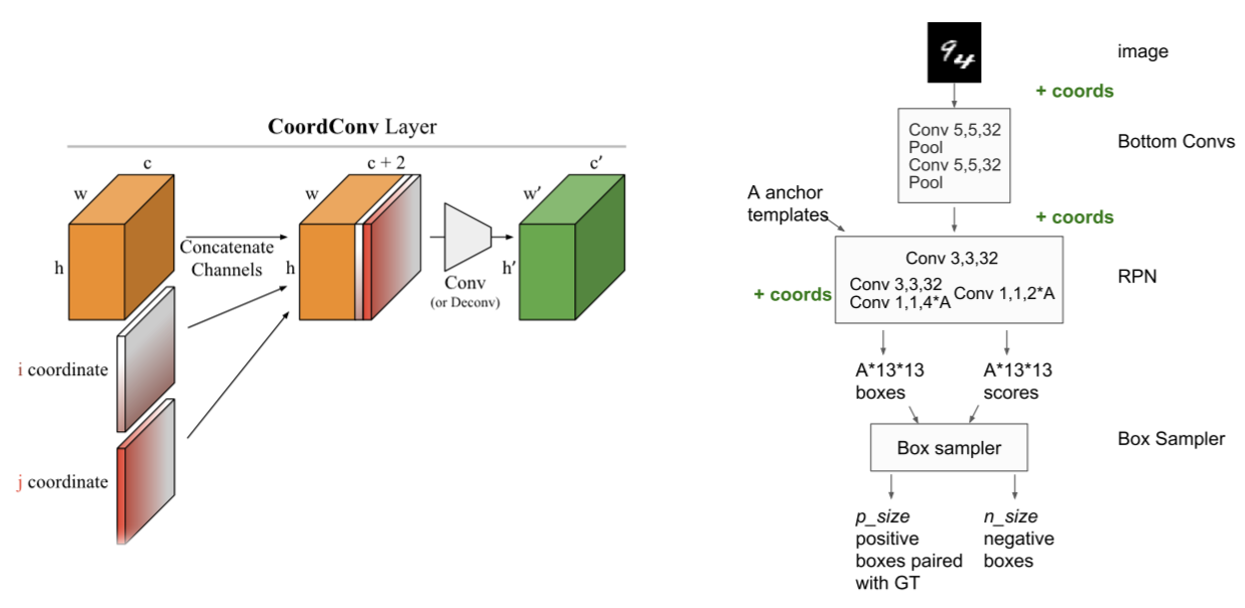}
\end{center}
   \caption{The structure of CoordConv layers (Left) and an object detection architecture with CoordConv (Right) proposed by Liu \etal \cite{liu2018intriguing}. }
\label{coordconv}
\end{figure*}

\section{Related work}

\noindent \textbf{CNN models for traffic sign detection.} Traffic sign detection is slightly different from common object detection scenarios. For simplicity, the frame of traffic sign is always in fundamental shapes and basic colors, such as red cycle, black rectangle and so forth. Meanwhile, central pictograph represents meaning of traffic signs. But for complexity, bad physical conditions like shadow, deformation and rust in reality are expected to be a major challenge in the detection process. Thus, detectors that  excel in recognizing real-time transformation with high detection accuracy are in demand. 

Generally, current CNN models for German Traffic Sign Detection Benchmark (GTSDB) \cite{arcos2018evaluation} are based on the various state-of-the-art object detection models, such as Faster R-CNN, R-FCN, SSD, and YOLO. They first initialize the model parameters pre-trained on the COCO dataset \cite{coco}, and then fine-tune them on the traffic sign detection datasets with backbone image feature extractors like Resnet V1 101, Inception V2, Inception Resnet V2, and Mobilenet V1. Finally,  take key metrics, including mean average precision (mAP) and execution time for a batch size of one, to evaluate and compare outcomes over four categorised traffic sign detection. As selected results shown in \Cref{cnns}, modules do not balance very well between mAP and running time.

\begin{table}[h]
\begin{center}
\begin{tabular}{l|c|c}
\hline
Model & mAP & Latency \\
\hline\hline
Faster R-CNN Inception Resnet V2 & 98.77 & 500 ms\\
YOLO V2 & 78.33 & 21.48 ms\\
SSD Inception V2 & 66.10 & 15.14 ms\\
\hline
\end{tabular}
\end{center}
\caption{The mAP and latency measurement on the simplified traffic sign categories as being a) Prohibitory, b) Mandatory, c) Danger, or d) Other, for various kinds of object detection models on the GTSDB test set.}
\label{cnns}
\end{table}


\noindent \textbf{Failing of CNN.} 
A simple classification and regression task conducted by Liu \etal \cite{liu2018intriguing} clearly illustrates the defect of CNN towards coordinate transform. As \Cref{fail} shown, with the goal of highlighting one-hot pixels via learning in Cartesian coordinate space, it turned out CNN performed well over training set in spite of uncertain noises emerging. However, prediction over testing set receive a large fail to generalize highlight pixels. Likewise, the similar outcomes happened in regression task. Thus far, the test argued that CNN had much difficulties dealing with coordinate transform problem. 

One solution that has been proposed is to replace the convolutional layer with the CoordConv layer \cite{liu2018intriguing}. As shown in \Cref{coordconv}, two more input coordinate channels are give to convolution layer, ensuring CNN to learn complete coordinate information. In practical operation, $i$ and $j$ coordinates layers are simply filled with row numbers or coulum numbers. Since extra channels are added to the original features, this approach introduces new kernel parameters in the following convolutional layers. The right part of \Cref{coordconv} shows that these extra channels should be added throughout the model, and thus this approach can not be applied to the pretrained CNN models that use the traditional convolutional layers.

\section{Coordinate Embedding - A Simple Fix}

\begin{figure}[htb]
\begin{center}
\includegraphics[width=0.8\linewidth]{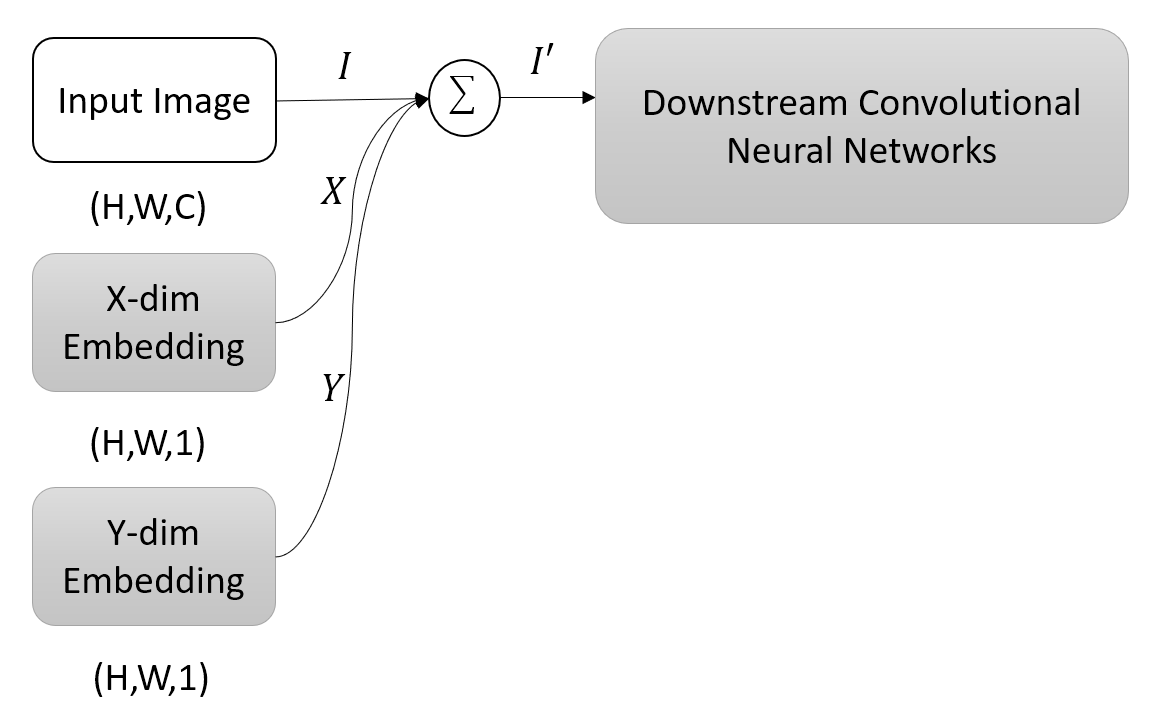}
   \end{center}
   \caption{The structure of the Coordinate Embedding layer.}
\label{ce}
\end{figure}

As shown in \Cref{ce}, we construct the Coordinate Embedding (CoordEmb) using two trainable matrices $X\in R^{H\times W\times 1}, Y \in R^{H\times W\times 1}$ to embed the x coordinate and y coordinate information of the input image $ I \in R^{H\times W\times C}$ respectively, where $H$ is the height of the input image, $W$ is the image width and $C$ represents the number of the image channels. Since the original input image is normalized to be in the range $I_{i,j,k}\in [-1,1]$, we also require the initial value of the x-dim embedding and y-dim embedding be normalized in the range $X_{i,j,k} \in [-1,1], Y_{i,j,k} \in [-1,1]$. The final output of the coordinate embedding layer, $I'$, is normalized to maintain the data range of the original image,

$$
I'=(I+X+Y)/3
$$

\begin{figure}[htb]
\begin{center}
\includegraphics[width=0.8\linewidth]{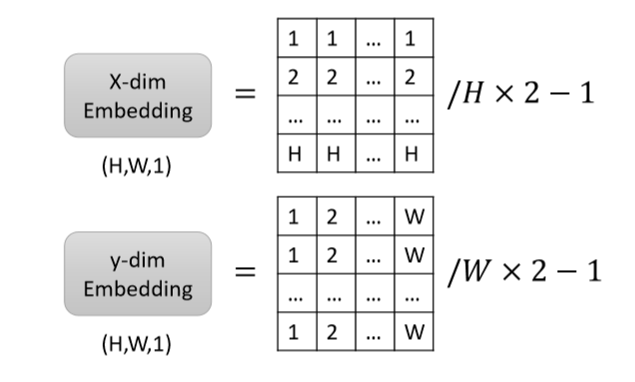}
\end{center}
   \caption{The initialization for the x-dim embedding and the y-dim embedding}
\label{coor}
\end{figure}

The coordinate information is introduced through an initialization of the x-dim and y-dim embeddings with their normalized x and y coordinates separately, as depicted in \Cref{coor}.

Our approach only introduced new parameters at the very beginning of the original CNN model with the number of the 2/3 size of the input image. Thus it can be applied to any downstream CNN models for any two dimensional computer vision tasks, and our approach can be simply extended to the three dimensional tasks by adding a new z-dimensional embedding matrix. The downside of our approach is that the model parameters will then be proportional to size of the input image and the translation invariance is harmed by introducing the positional information. However, considering the fact that human eyes have a fixed number of receptors on the retina, the scale of the vision signals into human retina are actually fixed sized images and those receptors may have different activation bias due to their positions on the human retina, and this kind of activation bias is what actually modeled by our coordinate embedding matrix. This kind of positional activation bias of the retina receptors is important and can be exemplified with the fact that human is more sensible to the moving objects near the edges of the Field of View (FoV) than the moving objects that happens in the center of FoV, which helps the ancient human beings to be more easily alerted for the incoming dangers and survive. 

As for the harm of the translation invariance possessed by the vanilla CNN, we believe that any kinds of invariance should be learned as model parameters through the interaction between the model and the environment. Given the fact that affine invariance is not favorable in all conditions of vision cognition (e.g. a common visual illusion image that can be viewed as duck in one rotation angle and also viewed as a rabbit in the other angle), we in turn hypothesize that the invariance property should be as the learned model parameters that can be applied locally on the input image but not as the structure priors through the design of the model architecture that applies globally to the whole input image.

\section{Experimentation}
The sections below describe the dataset setup and the specific architecture of the downstream CNN model.  
\subsection{Datasets}
We evaluate Coordinate embedding CNN using German Traffic Sign Detection Benchmark (GTSDB) \cite{houben2013detection}, which has been highly accepted and typically used for studies in traffic sign detection field. This dataset contains 900 full images including 1206 traffic signs from 43 categories. 600 images with 846 traffic signs are arranged for training purpose and the rest 300 images with 360 traffic signs for testing consideration. Arcos (2018) \etal \cite{arcos2018evaluation} categorised traffic sign as being a) Prohibitory, b) Mandatory, c) Danger, or d) Other, and detailed sign labels were discarded in their analysis. Note that the initiative wins in mAP as well as execution time, but four categories is less instructive in practical application like an autonomous driving or driving assistance system. Thus in our experiments, we use the full 43 categories for both model training and evaluation. 

\begin{figure}[htb]
\begin{center}
\includegraphics[width=0.8\linewidth]{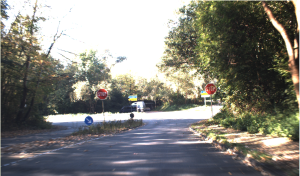}
\end{center}
   \caption{A sample image from GTSDB}
\label{fig:long}
\label{fig:onecol}
\end{figure}

\subsection{Single Shot Detector (SSD)}
 In this project, the official Tensorflow implementation \footnote{\url{https://github.com/tensorflow/models/blob/master/research/object_detection/models/ssd_inception_v2_feature_extractor.py}} of the Single Shot Detector with Inception V2 backbone (SSD Inception V2)\cite{ssd} is applied as the downstream CNN model.  Specifically, for a given input image with ground truth labels, SSD will do the following:

Firstly, transfer the image through a series of convolutional layers, producing several sets of feature maps at different scales.

Secondly, for each location in each of these feature maps, use a 3x3 convolutional filter to evaluate a small set of default bounding boxes. 

Thirdly, for each box, simultaneously predict: a) the bounding box offset, and b) the class probabilities

Lastly, during training, match the ground truth box with these predicted boxes based on Jaccard index. The best predicted box will be labeled a “positive,” along with all other boxes that have an Jaccard index with the truth $> 0.5$.

Our downstream CNN model is initialized with the parameters pre-trained on the COCO dataset, and the whole model including the CoordEmb layer is fine-tuned on the GTSDB dataset with the RMSprop optimizer using a learning rate of 0.004 for 155k training steps with a batch size of 24. 

\section{Results}

\begin{figure}[htb]
\begin{center}
\includegraphics[width=1.0 \linewidth]{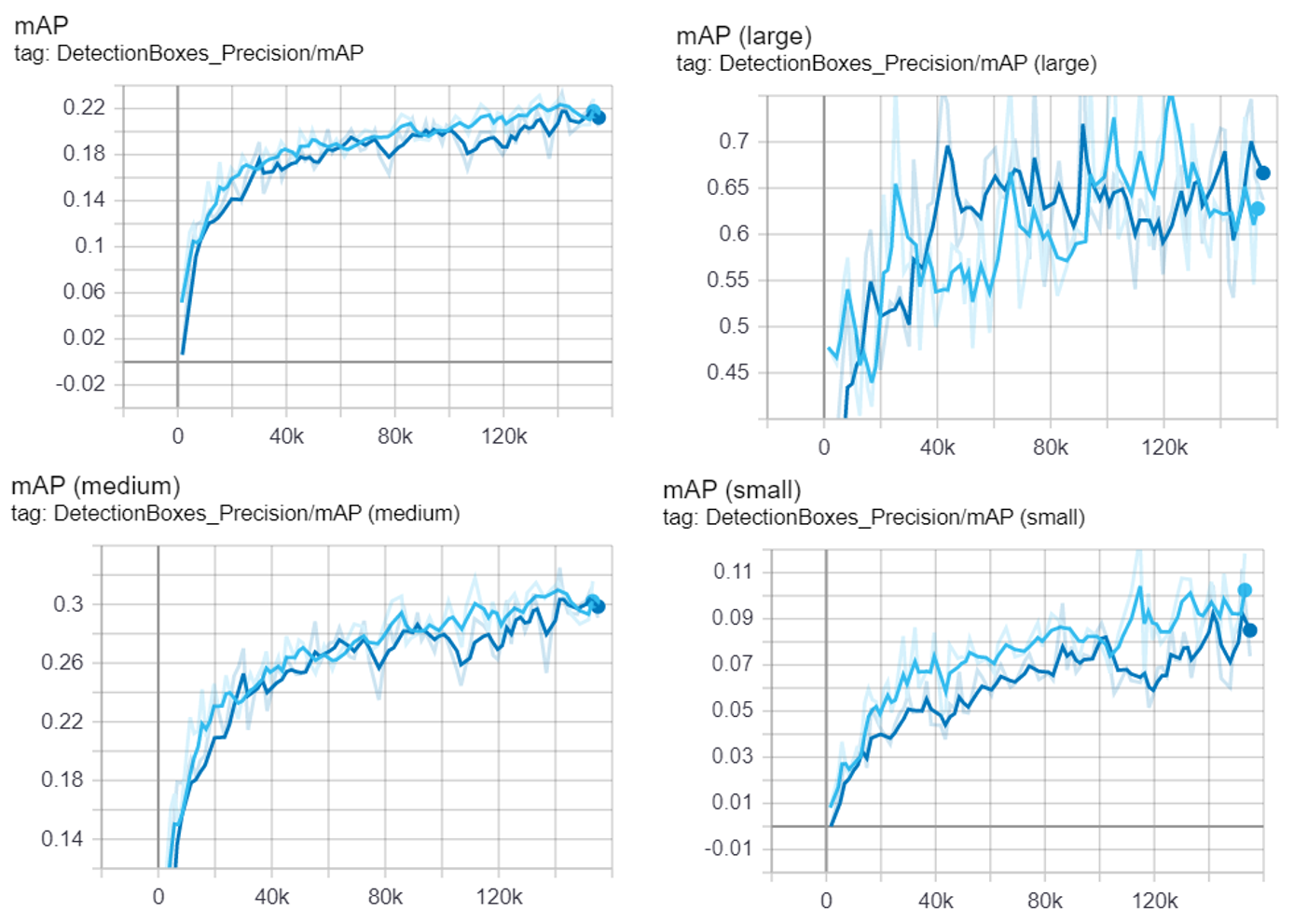}
\end{center}
   \caption{The mAP scores on the test set of GTSDB with respect to the training steps for the vanilla SSD Inception V2 (in dark blue) model and the CoordEmb+SSD Inception V2 model (in light blue).}
\label{res}

\end{figure}

\Cref{res} shows the mAP scores on the test set of GTSDB with respect to the training steps for the vanilla SSD Inception V2 (in dark blue) model and the SSD Inception V2 model with CoordEmb layer added at the very beginning stage of the input image (in light blue). We can observe that the model with our proposed CoordEmb consistently outperforms the vanilla model regarding to the metric of mAP throughout the training steps. The final converged mAP of the baseline model is 0.2041, while it is 0.2288 for the model with the CoordEmb, which means that our approach can bring a significant 2.47\% absolute mAP performance boost compared with the vanilla SSD Inception V2 model. 

From \Cref{res}, we can also see that our model is especially good at detecting the small and medium sized objects, while have a similar performance on the large sized objects compared with the vanilla model. This is because the small traffic signs often exists at the sides of the road that is far in distance and thus is often near the edges of the image. Our coordinate embeddings can capture this kind of relation between the positions and the objects and boosts the sensitivity of the model detections with respect to the small traffic signs.

\begin{figure*}[htb]
\centering  
\subfigure[SSD Inceptino V2]{
\label{Fig.sub.1}
\includegraphics[width=0.42\textwidth]{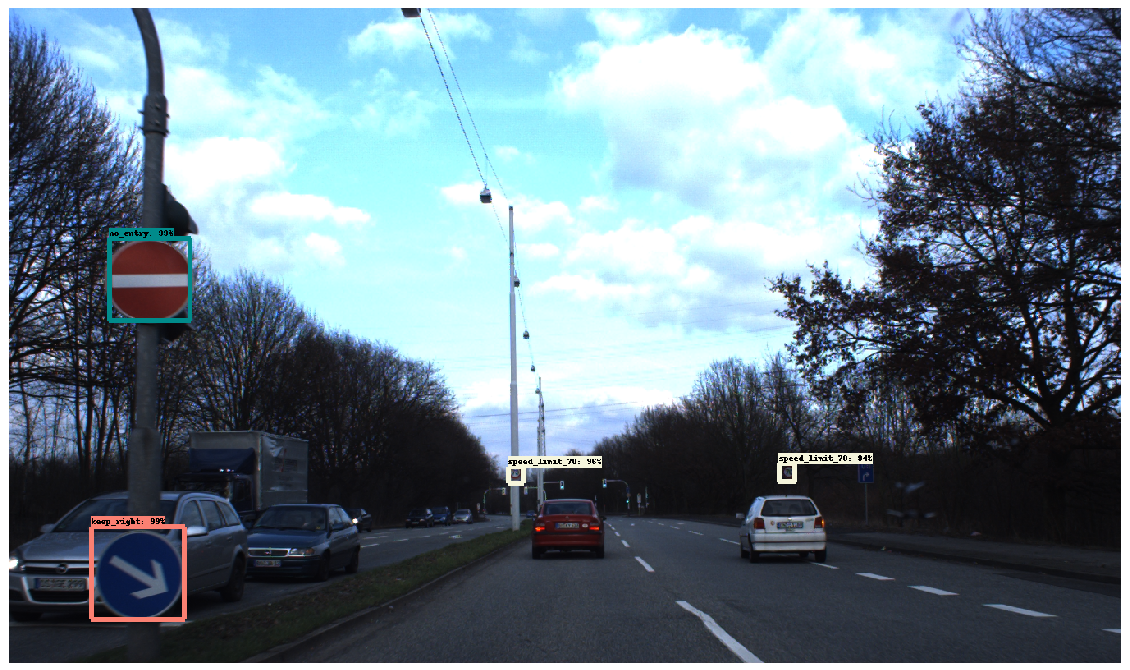}}
\subfigure[CoordEmb+SSD Inception V2]{
\label{Fig.sub.2}
\includegraphics[width=0.42\textwidth]{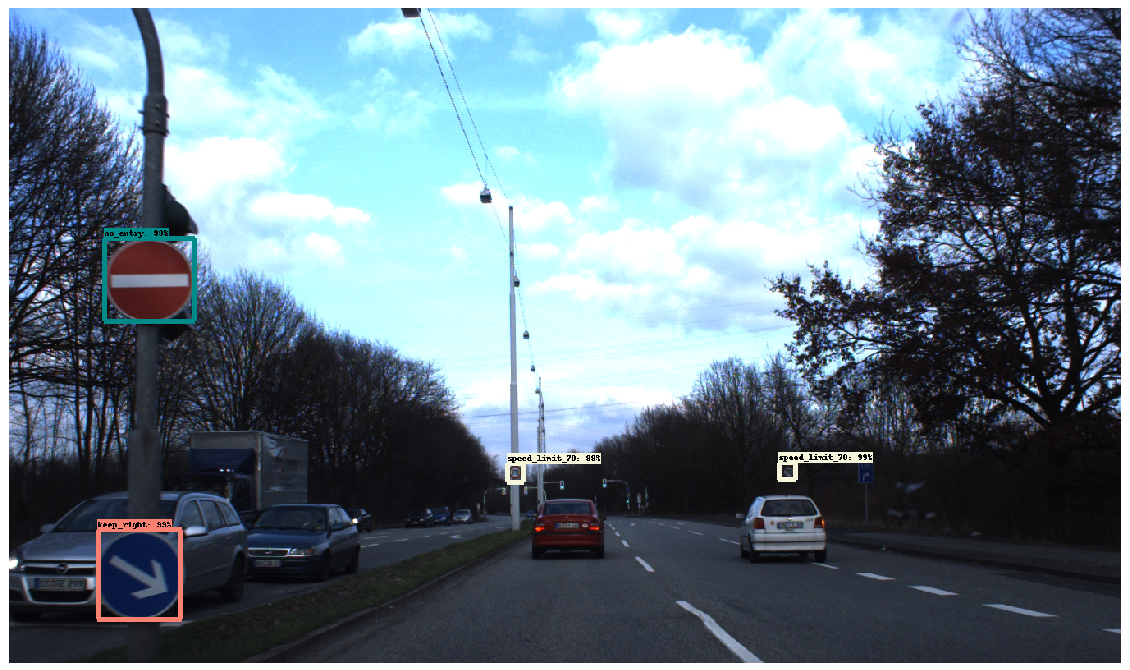}}
\caption{GTSDB sample detection results}
\label{Sample}
\end{figure*}

\begin{figure*}[htb]
\centering 
\subfigure[SSD Inception V2]{
\label{Fig.sub.1}
\includegraphics[width=0.42\textwidth]{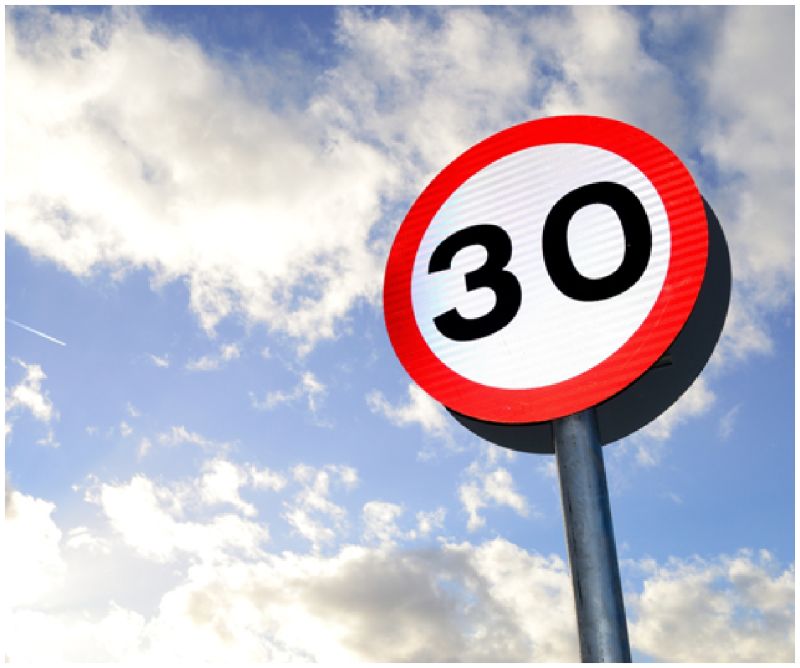}}
\subfigure[CoordEmb+SSD Inception V2]{
\label{Fig.sub.2}
\includegraphics[width=0.42\textwidth]{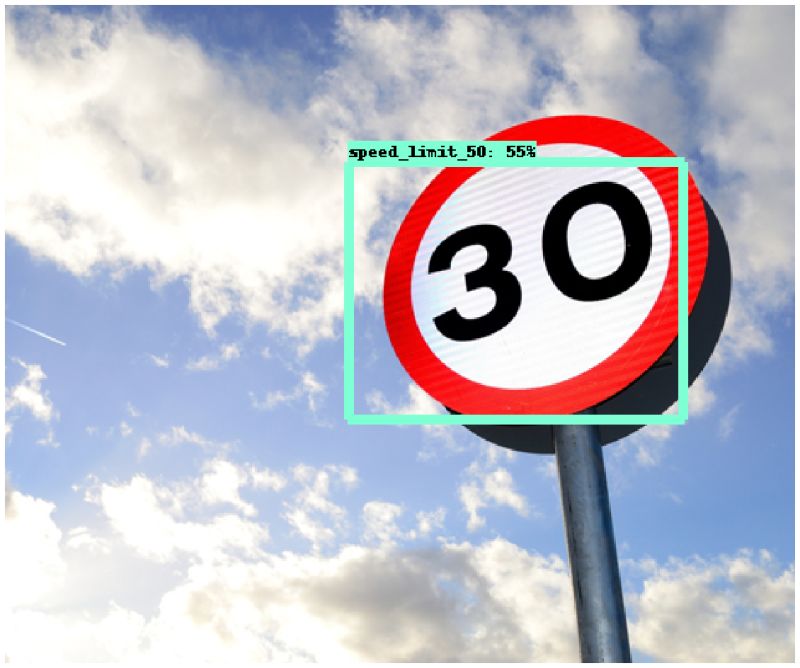}}
\subfigure[SSD Inception V2]{
\label{Fig.sub.3}
\includegraphics[width=0.42\textwidth]{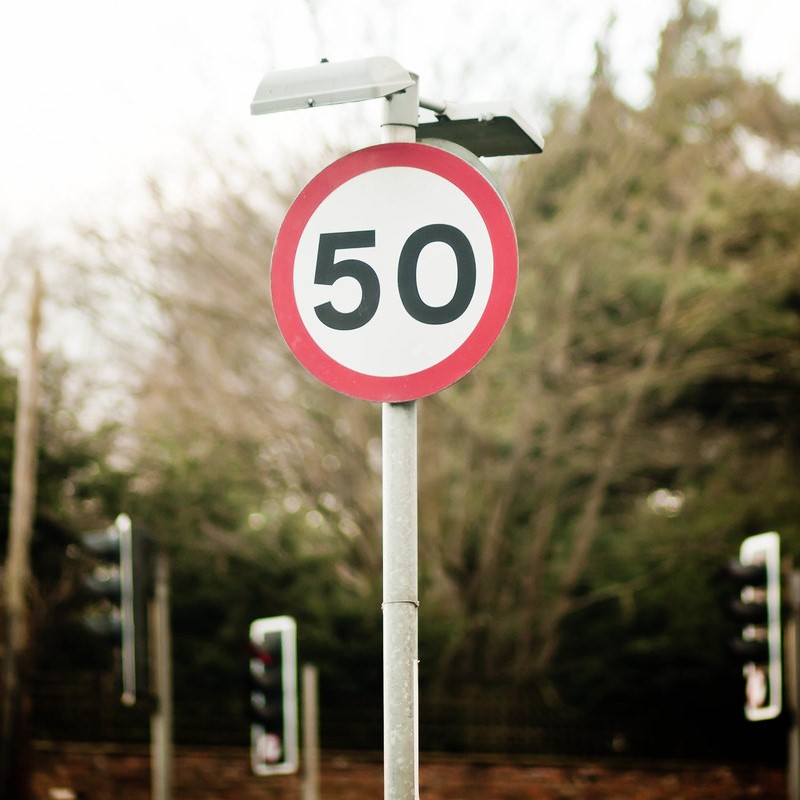}}
\subfigure[CoordEmb+SSD Inception V2]{
\label{Fig.sub.4}
\includegraphics[width=0.42\textwidth]{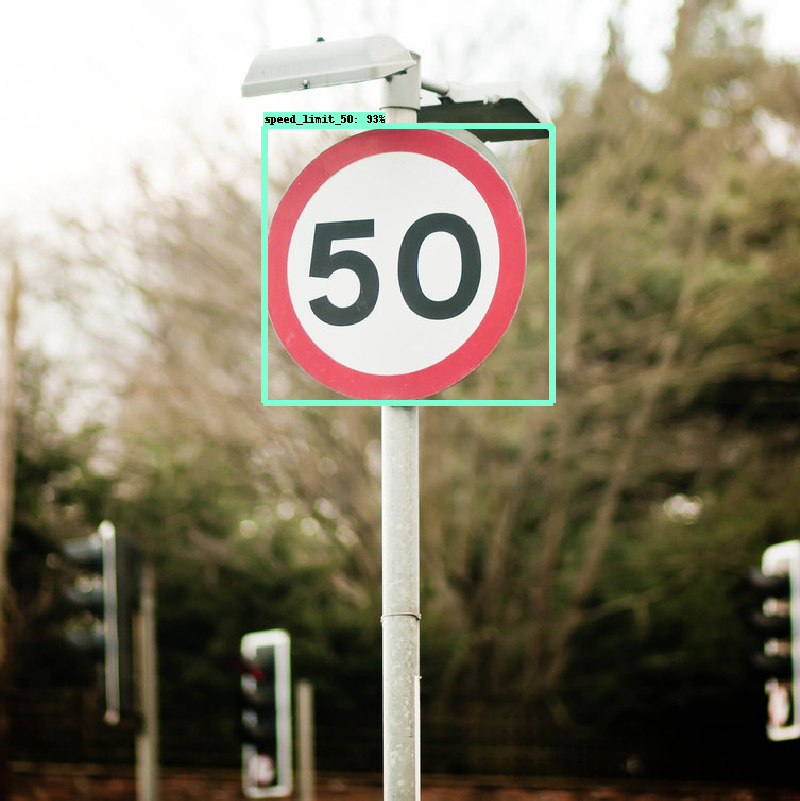}}
\caption{The object detection results for the affine distorted images}
\label{aff}
\end{figure*}

\section{Qualitative Analysis}

To consider the comprehensive performance of proposed model in real world detection scenarios, we qualitatively compare the model performance on both the images from the test set of GTSDB and the images with the affine distortion.  

\begin{table}[htb]
\begin{center}
\begin{tabular}{l|c|c}
\hline
Label & Original CNN & CoordEmb CNN \\
\hline\hline
No entry & 99\% & 99\%\\
Keep right & 99\% & 99\%\\
Speed limit 70 (left) & 96\% & 99\%\\
Speed limit 70 (right) & 94\% & 99\%\\
\hline
\end{tabular}
\end{center}
\caption{the Confidence score of the model }
\label{accu}
\end{table}

\Cref{Sample} and \Cref{accu} reveal that CoordEmb model has equal share with vanilla model in regard of traffic signs large size, i.e., No entry, Keep right, but holds much more confidence with respect to small and marginal traffic signs, such as two speed limit at the far end, increasing accuracy in magnitude of 3\% and 5\% respectively. Besides, proposed model is significantly superior when slight coordinate transformations happen on detected object. \Cref{aff} shows two special cases in detection, where traffic signs are not well-placed from the perspective of detector, our model is still robust with acceptable reliability of 55\% and 93\%. In contrast, vanilla model has assigned them to unrecognizable cases. The outcomes remarkably corroborate the model's capability of sensitive detection in reality.

\section{Conclusion and future work}
In this paper, we proposed a novel Coordinate Embedding (CoordEmb) approach that introducing the positional information to CNN models with the following advantages: (1) it adds minimal extra computations and parameters to the original model. (2) it can be applied to the existing pretrained CNN models without model architecture modification. (3) the model with CoordEmb significantly beats the vanilla CNN model on the traffic sign detection task. (4) it enables the model with CoordEmb to be more robustness with respect to the affine distortion compared with the vanilla CNN model.

Our future work is to test our CoordEmb approach on a larger scale of dataset like COCO\cite{coco} to see if it can boost the performance of the state-of-the-art models, and to compare the CoordEmb model trained from scratch with the CoordConv approach to see if our approach can give better results apart from the fine-tuning paradigm.

{\small
\bibliographystyle{ieee_fullname}
\bibliography{egpaper_for_review}
}

\end{document}